%% file: main_neurips.tex
\documentclass{article}

\usepackage[final]{neurips_2024}

\usepackage[utf8]{inputenc} 
\usepackage[T1]{fontenc}    
\usepackage{hyperref}       
\usepackage{url}            
\usepackage{booktabs}       
\usepackage{amsfonts}       
\usepackage{nicefrac}       
\usepackage{microtype}      
\usepackage{xcolor}         
\usepackage{graphicx}
\usepackage{multirow} 
\usepackage{makecell}
\newcommand{\myparagraph}[1]{{\noindent\bf #1}}
\usepackage{wrapfig}
\usepackage{amsmath}
\usepackage{natbib}
\setcitestyle{numbers,square}
\usepackage[linesnumbered,ruled,vlined]{algorithm2e}

\usepackage[misc]{ifsym} 
\newcommand\blfootnote[1]{%
    \begingroup
    \renewcommand\thefootnote{}
    \footnotetext{#1}
    \addtocounter{footnote}{-1}
    \endgroup
}

\newcommand{\xg}[1]{\textcolor{blue}{#1}}

\title{Video Diffusion Models are Training-free \\
Motion Interpreter and Controller}

\author{%
  Zeqi Xiao$^{1}$, Yifan Zhou$^{1}$, Shuai Yang$^{2}$, Xingang Pan$^{1}$\ \\
  $^{1}$S-Lab, Nanyang Technological University, \\
  $^{2}$Wangxuan Institute of Computer Technology, Peking University \\
  \texttt{\{zeqi001, yifan006\}@e.ntu.edu.sg} \\
  \texttt{williamyang@pku.edu.cn, xingang.pan@ntu.edu.sg}
}

\begin{document}

\maketitle

\blfootnote{Project page at this \href{https://xizaoqu.github.io/moft/}{URL}.}

\input{sections/abstract}
\input{sections/introduction}
\input{sections/related_work}

\input{sections/preliminary}
\input{sections/method}

\input{sections/experiments}
\input{sections/conclusion}

\clearpage
\bibliographystyle{splncs04}
\bibliography{main}

\newpage

\input{sections/supp}

\end{document}

%% file: sections/abstract.tex
\begin{abstract}
Video generation primarily aims to model authentic and customized motion across frames, making understanding and controlling the motion a crucial topic. 
Most diffusion-based studies on video motion focus on motion customization with training-based paradigms, which, however, demands substantial training resources and necessitates retraining for diverse models.
Crucially, these approaches do not explore how video diffusion models encode cross-frame motion information in their features, lacking interpretability and transparency in their effectiveness.
To answer this question, this paper introduces a novel perspective to \textit{understand}, \textit{localize}, and \textit{manipulate} motion-aware features in video diffusion models. Through analysis using Principal Component Analysis (PCA), our work discloses that robust motion-aware feature already exists in video diffusion models. We present a new MOtion FeaTure (MOFT) by eliminating content correlation information and filtering motion channels. MOFT provides a distinct set of benefits, including the ability to encode comprehensive motion information with clear interpretability, extraction without the need for training, and generalizability across diverse architectures. Leveraging MOFT, we propose a novel training-free video motion control framework. Our method demonstrates competitive performance in generating natural and faithful motion, providing architecture-agnostic insights and applicability in a variety of downstream tasks.

\begin{figure}[t]
    \centering
    \includegraphics[width=\linewidth]{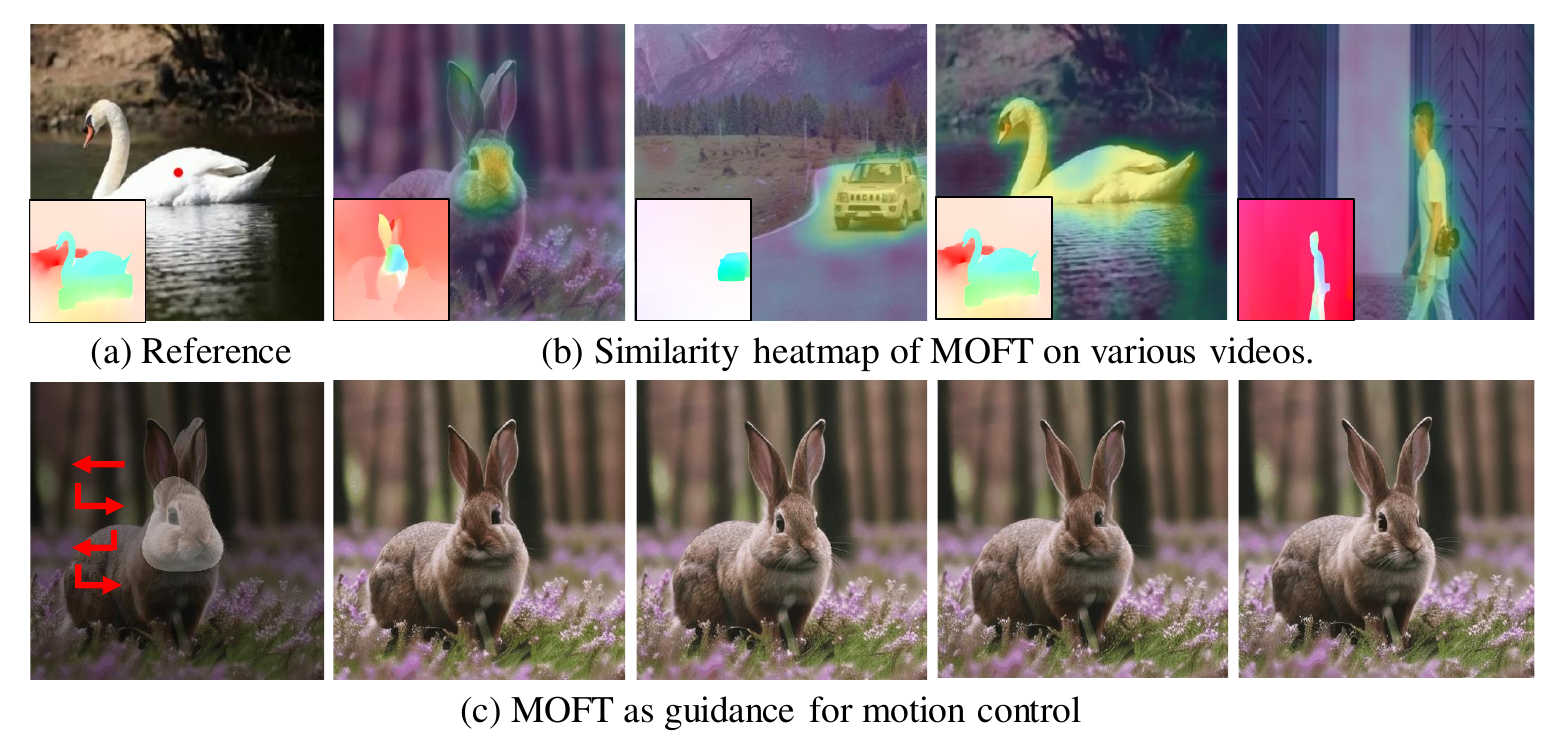}
    \caption{\textbf{Characteristics of MOtion FeaTure (MOFT).} (a-b) Rich Motion Information: We extract MOFT at the red point in the reference video in (a) and draw similarity heatmaps in (b) across various videos (yellow indicates higher similarity). The heatmap aligns well with the motion flow in the bottom left. (c) MOFT serves as guidance for controlling motion direction in the light-masked region, with the motion direction signal illustrated by red arrows in the first image.}\vspace{-1mm}
    \label{fig:teaser}
\end{figure}


\end{abstract}

%% file: sections/introduction.tex
\section{Introduction}

Video generation has experienced notable advancements in recent years, particularly in the realm of video diffusion models, such as text-to-video (T2V) generation \cite{guo2023animatediff, chen2023videocrafter1, wang2023lavie, wang2023modelscope} and image-to-video (I2V) generation \cite{blattmann2023stable, guo2023sparsectrl, chen2024videocrafter2}.
Apart from producing high-quality content in individual frames, capturing authentic and customized motion across frames is a crucial feature of video generation.
Thus, understanding and controlling the motion play pivotal roles in video generation.

Most methods \cite{wang2023motionctrl,yin2023dragnuwa,wang2024boximator,guo2023animatediff, wang2024videocomposer} that study video motion focus on motion customization, i.e. allowing users to specify a moving direction \cite{wang2023motionctrl} or a point-drag command \cite{yin2023dragnuwa}.
These methods typically adopt training-based paradigms \cite{wang2023motionctrl, yin2023dragnuwa, wang2024boximator, guo2023animatediff} that introduce motion conditions and train additional modules to ensure that the output videos adhere to these conditions. Despite their progress, these approaches require significant training resources and need retraining for different models, and their effectiveness often remains black-box. More critically, they do not address a fundamental question: \textit{How do video diffusion models encode cross-frame motion information within their features?}

Understanding the encoding of motion information is crucial for two reasons: a) it offers architecture-agnostic insights, meaning that such knowledge can be applied across different models and their checkpoints, an important consideration given the rapid evolution of video diffusion models; and b) it supports various downstream applications. For instance, the DIffusion FeaTure \cite{tang2024emergent} demonstrates how diffusion features can encapsulate rich semantic information, enabling applications like correspondence extraction \cite{hedlin2024unsupervised, ju2024robo} and image/video editing \cite{dhariwal2021diffusion, mou2023dragondiffusion,deng2023dragvideo}.

To this end, this paper introduces a novel perspective to \textit{understand}, \textit{localize}, and \textit{manipulate} motion-aware features in video diffusion models. We first establish that removing content correlation information helps to pronounce motion information in video diffusion features.
By applying Principal Component Analysis (PCA) \cite{wold1987principal} on these diffusion features, we observe a strong correlation between the principal components and video motions. Further explorations reveal that certain channels of the features play a more significant role in determining motion direction than others. Based on these observations, we present a straightforward strategy to extract motion information embedded in the features, termed MOtion FeaTure (MOFT). Through content correlation removal and motion channel filter, MOFT establishes impressive correspondence on videos with the same motion direction, as illustrated in Fig. \ref{fig:teaser} (a-b). Importantly, this strategy proves to be generalizable across various text-to-video or image-to-video generation models \cite{guo2023animatediff, guo2023sparsectrl, wang2023modelscope, zeroscope, blattmann2023stable} (Fig. \ref{fig:MOFT_visualize}), such as AnimatedDiff \cite{guo2023animatediff}, ModelScope \cite{wang2023modelscope}, and Stable Video Diffusion \cite{blattmann2023stable}.

Building upon the motion-aware MOFT, we propose a pipeline for video motion control in a training-free manner, without the modification of model parameters. The approach leverages compositional loss functions for content manipulation \cite{dhariwal2021diffusion, epstein2024diffusion, nichol2021glide, bansal2023universal,ho2022classifier}. Specifically, we design loss functions to optimize noisy latents in the denoising process with reference MOFT, which can be synthesized via direction signal or extracted from reference videos. 
Furthermore, our pipeline can be extended for point-drag manipulation. With MOFT guidance to generate coarse motion in the early denoising stages, fine-grained point-drag manipulation with DIFT \cite{tang2024emergent} guidance becomes feasible for videos. 
Various experiments showcase the effectiveness of MOFT in controlling the motions of diverse scenarios across different video diffusion models without the need for any training. Remarkably, our training-free method even outperforms some data-driven methods in achieving natural and faithful motion.
Our main contributions are summarized as follows:
\begin{itemize}
  \item We perform a deep analysis of motion information embedded in video generation models. Our work discloses that robust motion-aware feature already exists in video diffusion models.
  \item Through our analysis, we present MOtion FeaTure (MOFT) that effectively captures motion information. 
  MOFT has several advantages: 
  a) it encodes rich motion information with high interpretability;
  b) it can be extracted in a training-free way;
  and c) it is generalizable to various architectures.
  \item We propose a novel training-free video motion control framework based on MOFT. Our method demonstrates competitive performance with natural and faithful motion. Unlike previous training-based methods that need independent training for each different architecture and checkpoint, our method is readily applicable to different architectures and checkpoints.
\end{itemize}


%% file: sections/related_work.tex
\section{Related Works}

\myparagraph{Video Diffusion Models.}
The field of video generation has witnessed substantial progress in recent years, particularly in the domain of video diffusion models. Noteworthy contributions include advancements in text-to-video (T2V) generation \cite{ho2022video, guo2023animatediff, chen2023videocrafter1, wang2023lavie, wang2023modelscope, sora, ling2024motionclone, wu2024motionbooth, guo2024liveportrait} which aim to generate high-fidelity videos that align with textual descriptions. Besides, image-to-video (I2V) \cite{blattmann2023stable, guo2023sparsectrl, chen2024videocrafter2} takes image conditions as inputs and generates videos aligned with the image.
Beyond the production of high-quality content within individual frames, the capability to capture authentic and customized motion across frames stands out as a significant feature in the realm of video generation. 

\myparagraph{Diffusion Feature Understanding.}
The analysis and comprehension of diffusion features \cite{tang2024emergent, mo2023freecontrol, du2023generative, luo2023readout, si2023freeu} have garnered increasing attention. A comprehensive understanding of diffusion features not only yields architecture-agnostic insights applicable across diverse models and checkpoints but also enhances various downstream applications.
For instance, DIffusion FeaTure (DIFT) \cite{tang2024emergent} demonstrates that diffusion features embed impressive semantic correspondence and can be extracted with a simple strategy. This strategy proves effective across various architectures, spanning image diffusion models \cite{rombach2022high} to video diffusion models \cite{guo2023animatediff, wang2023modelscope}. Its versatility facilitates a range of applications, including correspondence extraction \cite{hedlin2024unsupervised, ju2024robo} and image/video editing \cite{dhariwal2021diffusion, mou2023dragondiffusion, deng2023dragvideo}. Recently, Freecontrol \cite{mo2023freecontrol} applied PCA on diffusion features and extracted semantic basics for training-free spatial control. Its method can be generalized to any conditional input and any model. Similarly, video diffusion models encode rich motion information within the features. However, less effort has been made to analyze it.


\myparagraph{Video Motion Control.}
Considerable efforts have been dedicated to tailoring video motion according to user preferences \cite{wang2023motionctrl,wang2024videocomposer,guo2023sparsectrl,yin2023dragnuwa,wang2024boximator,he2024cameractrl, chen2023livephoto, geyer2023tokenflow, yang2023rerender,wang2024humanvid, wang2024easycontrol, zhou2024trackgo,qiu2024freetraj,kansy2024reenact, zhang2024tora}. For example, MotionCtrl \cite{wang2023motionctrl} facilitates precise control over camera poses and object motion, allowing for fine-grained motion manipulation. VideoComposer \cite{wang2024videocomposer} introduces motion control through the incorporation of additional motion vectors, while DragNUWA \cite{yin2023dragnuwa} proposes a method for video generation that relies on an initial image, provided point trajectories, and text prompts.
These methodologies typically rely on training-based paradigms, incorporating motion conditions during training and requiring additional modules to ensure that the resulting videos adhere to these specified conditions. Despite their advancements, these approaches demand substantial training resources, necessitating retraining for different models, and often exhibit a black-box nature in terms of their effectiveness.
In contrast, this paper introduces a novel pipeline for controlling video motion using an interpretable motion-aware feature. Notably, this approach is training-free and can be generalized across various architectural frameworks, offering a more versatile and resource-efficient solution.

%% file: sections/method.tex
\section{MOtion FeaTures (MOFT)}


In this section, we first analyze how video diffusion models encode cross-frame motion information, then provide the strategy to extract motion features from pre-trained video diffusion models.

Similar to \cite{tang2024emergent,yatim2023space, mo2023freecontrol}, our analysis focuses on diffusion features extracted from the intermediate blocks of diffusion models. We denote them as $\mathcal{X} \in \mathbb{R}^{H \times W \times F \times D}$, where $H$, $W$, $F$ and $D$ are dimensions of height, width, frames, and channels, respectively.
As proved by prior works \cite{guo2023animatediff, zhao2023motiondirector, wang2023motionctrl}, cross-frame features play a crucial role in video motion control. For example, AnimateDiff \cite{guo2023animatediff} trains temporal self-attention LoRAs \cite{hu2021lora} that operate on the temporal dimension to control the global motion direction. Consequently, we argue that the temporal dimension encapsulates rich motion information. However, extracting motion information from diffusion features is non-trivial, as they also contain other information such as semantic and structural correlation.






\subsection{Content Correlation Removal}

Inspired by VideoFusion \cite{luo2023videofusion} which uses shared noise to model content correlation across frames and residual noise to model dynamic difference, we hypothesize that we can filter out the content correlation by eliminating similar information across frames:
\begin{equation}\label{eqn:correlation_removal}
    \mathcal{X}^{\text{norm}} = \mathcal{X} - \frac{1}{F}\sum_{i=1}^F \mathcal{X}_i,
\end{equation}
where $\mathcal{X}_i$ indicates the $i^{\text{th}}$ frame of feature $\mathcal{X}$. The shared latents, to which we refer as content correlation information, encompass shared aspects such as semantic content and appearance. In contrast, the residual latents primarily capture motion information, which also can be interpreted as deformation in structure.

\begin{figure}[t]
    \centering
    \includegraphics[width=\linewidth]{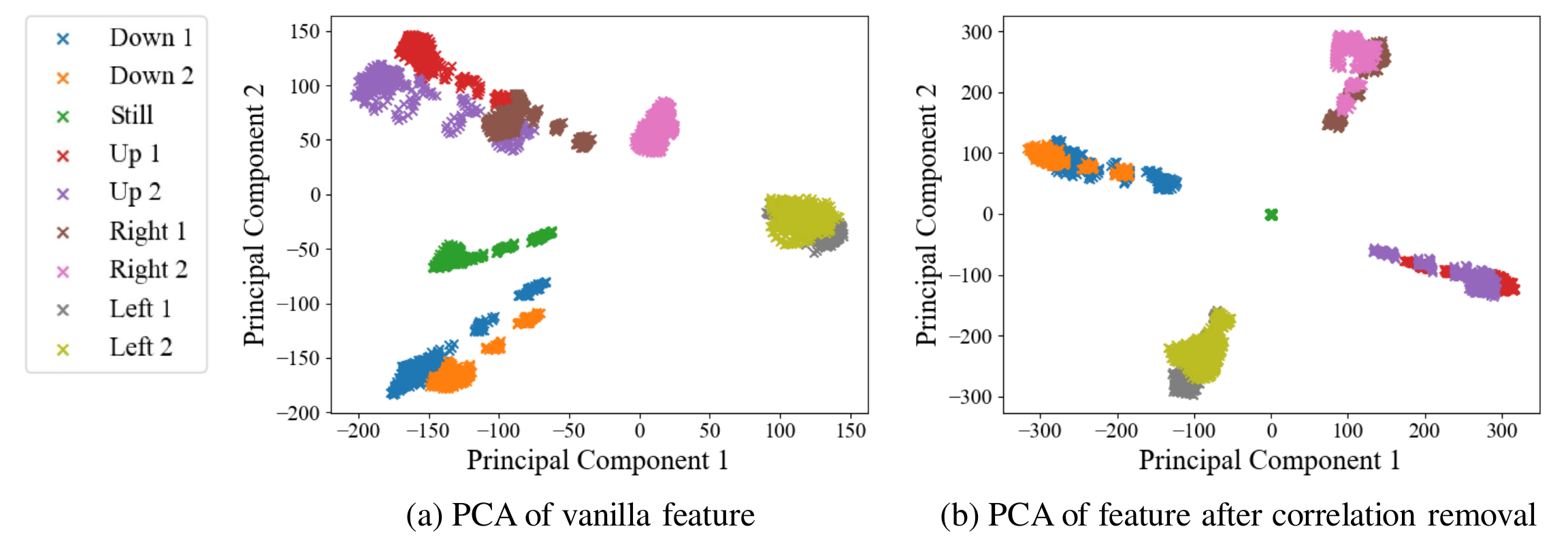}
    \vspace{-0.6cm}
    \caption{\textbf{Visualization of PCA on video diffusion features.} The left side indicates the frame-wise panning direction, with each color representing a specific direction pattern. We apply PCA to diffusion features extracted from videos with different motion directions and plot their projections on the leading two principle components. (a) The result does not exhibit a distinguishable correlation with motion direction. (b) Features are clearly separated by their motion direction.}
    \vspace{-0.2cm}
    \label{fig:pca}
\end{figure}

To validate the hypothesis, we apply Principal Component Analysis (PCA) \cite{wold1987principal} on $\mathcal{X}$ and $\mathcal{X}^{\text{norm}}$.
Specifically, we create a series of videos with the entire scene moving horizontally or vertically, resulting in a set of features $\{\mathcal{X}^1, \mathcal{X}^2, ..., \mathcal{X}^n\}$ extracted from videos in the process of DDIM \cite{song2020denoising} inversion. In this subsection, we omit the choice of video model architecture and feature selection for simplicity. 
We analyze and project the 
$D$-dimensional features of the first frame on the leading two principal components ($\mathcal{P}_1$ and $\mathcal{P}_2$). As shown in Fig. \ref{fig:pca} (a), the result of the vanilla feature does not exhibit a distinguishable correlation with motion direction. In contrast, as shown in Fig. \ref{fig:pca} (b), normalized features are successfully separated by their motion direction. It reveals that the normalization operation removes content correlation information and emphasizes motion information.


\begin{figure}[t]
    \centering
    \includegraphics[width=\linewidth]{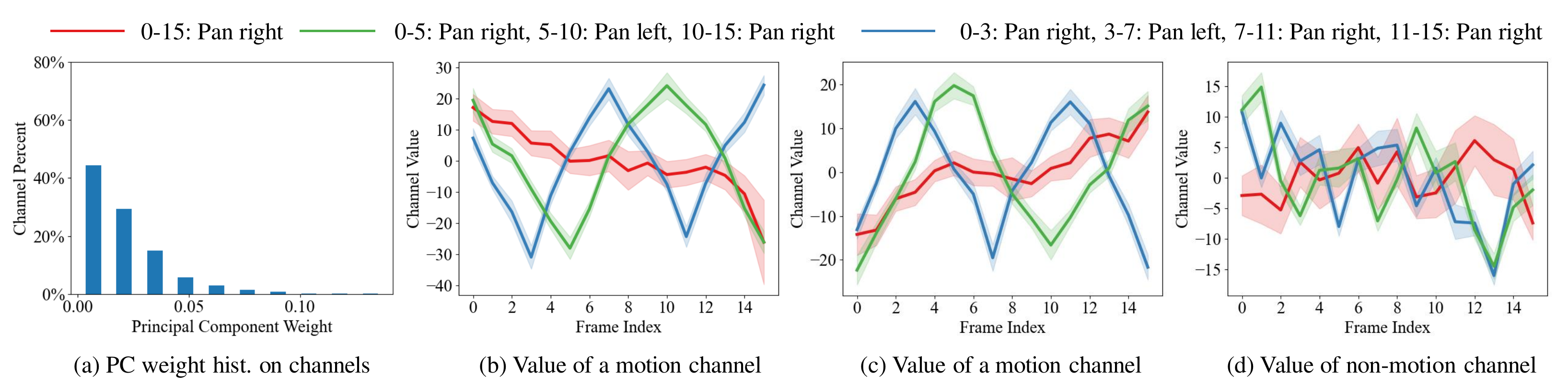}
    \vspace{-12pt}
    \caption{\textbf{Cross-frame Channel Value.} (a) We plot the histogram of the weight of $\mathcal{P}_1$. It reveals that only a few channels significantly contribute to determining the principal components. (b-c) The motion channels exhibit a pronounced correlation with motion direction trends. (d) In contrast, the non-motion channels show little correspondence with motion direction.}\vspace{-0.3cm}
    \label{fig:channel_line_chart}
\end{figure}

\subsection{Motion Channel Filter}

Principal components can not only reduce dimension but also reflect the importance of each dimension by the projection weights.
We visualize the projection weights of $\mathcal{P}_1\in\mathbb{R}^{D\times1}$ in Fig \ref{fig:channel_line_chart} (a). It reveals that only a few channels significantly contribute to determining the principal components, indicating these channels encode richer motion information. We term them Motion Channels.


To further explore the relationship between these channels and the motion in videos, we create videos panning in different directions at various frames and visualize the channel with the highest two projection weights in $\mathcal{P}_1$. As depicted in Fig. \ref{fig:channel_line_chart} (b-c), the value trend is closely associated with the panning direction of the video. Specifically, in Fig. \ref{fig:channel_line_chart} (b), the motion channel value decreases during a rightward pan and increases during a leftward pan. In contrast, a channel with low projection weight does not exhibit much correspondence (Fig. \ref{fig:channel_line_chart} (d)).
These observations indicate that we can extract motion-aware features by filtering these motion channels.


\begin{figure}[t]
    \centering
    \includegraphics[width=\linewidth]{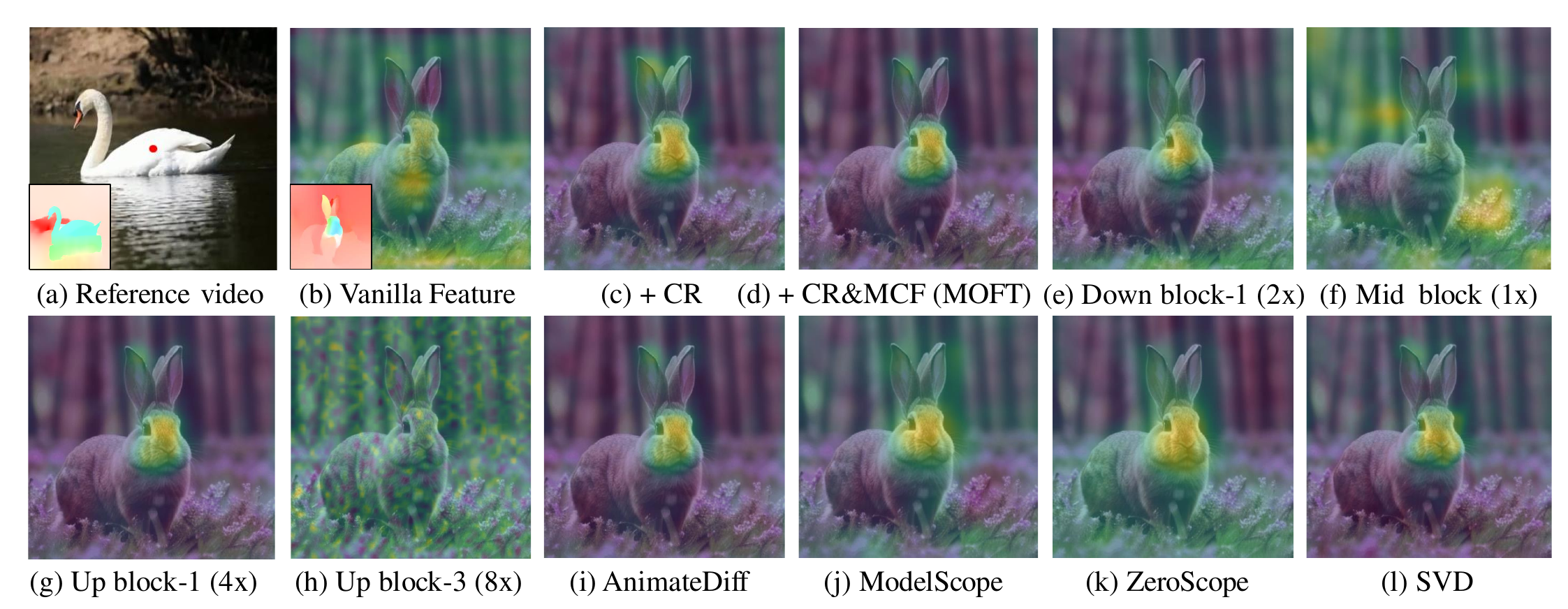}
    \vspace{-12pt}
    \caption{\textbf{Similarity heatmap between feature of the source point and target features}. Given the red source point in (a), we plot the similarity heatmap on target videos. Yellow indicates regions with higher similarity. We normalize all similarity to 0-1 for better illustration. (b-d) Similarity heatmap of features with different designs. ``CR'' indicates ``content removal''. ``MCF'' indicates motion channel filter. (e-h) Similarity heatmap of MOFT in different layers in the U-Net. (2x) means relative spatial resolution scale 2. (i-l) Similarity heatmap of MOFT in different video generation models.}\vspace{-3mm}
    \label{fig:MOFT_visualize}
\end{figure}

\subsection{MOFT Extraction}

With the above explorations, we introduce a straightforward strategy for extracting motion information from video diffusion models, which we term Motion Feature (MOFT). Our method includes two designs: content correlation removal and motion channel filter. The process can be represented as follows
\begin{equation}
\mathcal{M} = (\mathcal{X}_{[j]} - \frac{1}{F} \sum_{i=1}^{F} \mathcal{X}_{i,[j]}), \quad j \in \mathcal{C},
\label{alg:moft_extraction}
\end{equation}
where $\mathcal{M}$ is the extracted MOFT, $i$ operates on the frame dimension, and $j$ operates on channel dimension. $\mathcal{C}$ is the channel index set of motion channels. 


We illustrate how content correlation removal and motion channel filter improve the motion correspondence in Fig. \ref{fig:MOFT_visualize} (a-d). Vanilla video features demonstrate weak alignment with the reference motion. The proposed content correlation removal significantly improves the alignment. Further application of the motion channel filter enhances focus on the motion area (\textit{e.g.}, the rabbit head), yielding higher correspondence.



We conduct an additional ablation study and visualize the impact of selecting different video diffusion features from various blocks within the U-Net of AnimateDiff \cite{guo2023animatediff}. Fig. \ref{fig:MOFT_visualize} (e-h) intuitively reveals that features with relative medium resolutions achieve better motion correspondence. To this end, we select the features after upper block 1.

While the above analysis is based on AnimateDiff \cite{guo2023animatediff}, the property of MOFT holds in different base video models \cite{guo2023animatediff, wang2023modelscope, zeroscope, blattmann2023stable} (Fig. \ref{fig:MOFT_visualize} (i-l)), demonstrating that MOFT is versatile across different video generation frameworks, consistently achieving reliable motion alignment.


While MOFT is reminiscent of optical flow, which also describes motion, a key limitation of optical flow is that it cannot be directly extracted from video diffusion models during the denoising process and hence cannot serve as the guidance for motion control.
In contrast, MOFT is available even at early denoising steps and is naturally suitable for motion control, as we will discuss in the next section.

\section{MOFT Guidance}
With the motion-aware MOFT, we propose a pipeline for video motion control in a training-free manner (Sec. \ref{sec:motion_control}). Furthermore, our pipeline can be extended for point-drag manipulation (Sec. \ref{point_drag_manipualtion}).
\subsection{Motion Control}\label{sec:motion_control}

\begin{figure}[t!]
    \centering
    \includegraphics[width=\linewidth]{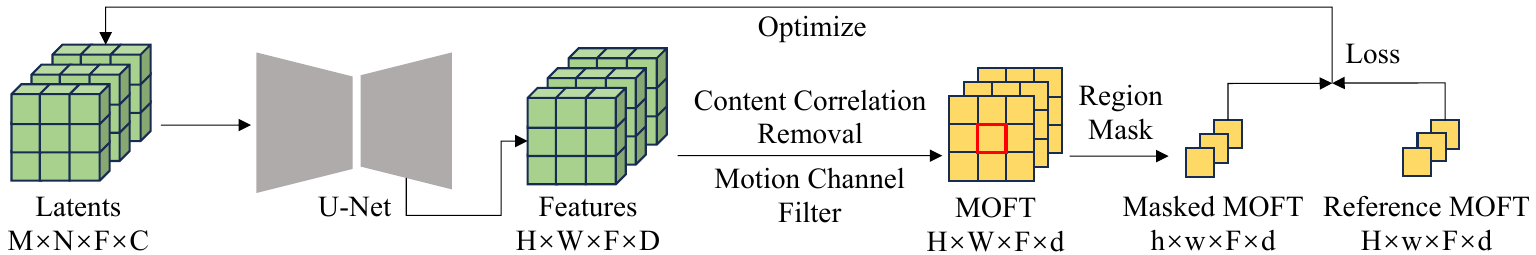}
    \vspace{-16pt}
    \caption{\textbf{Motion Control Pipeline}. We use reference MOFT as guidance and optimize latents to alter the sampling process. In one denoising step, we get the intermediate features and extract MOFT from it with content correlation removal and motion channel filter. We optimize the latents to alter the sampling process with the loss of masked MOFT and reference MOFT.}
    \vspace{-11pt}
    \label{fig:mainfig}
\end{figure}

We design a pipeline to control motion in the generated videos in a training-free way, as depicted in Fig.~\ref{fig:mainfig}. 
Following \cite{shi2023dragdiffusion,yatim2023space}, we optimize latents to alter the sampling process. The loss function $\mathcal{L}^c$ is
\begin{equation}
    \mathcal{L}^c = \frac{1}{|\mathcal{R}|} \sum_{(i,j)\in \mathcal{R}} || \mathcal{M}_{i,j} - \mathcal{M}^r_{i,j} ||,
\label{equ:motion_control_guidance}
\end{equation}
where $\mathcal{M}$ is the MOFT we extract during the denoising phase, $\mathcal{M}^r$ is the reference MOFT feature, and $\mathcal{R}$ is the position set of the region that we want to control motion.
We provide two possible ways to construct the reference MOFT $\mathcal{M}^r$: 1) Extract MOFT from reference videos. We perform DDIM inversion~\cite{song2020denoising} on reference videos and extract MOFT in the inversion stage. 
2) Synthesize MOFT based on the statistic regularity. As shown in Fig. \ref{fig:channel_line_chart} (b-c), frame-wise motion channel values exhibit high correspondence with frame-wise motion. We can fit it into a piecewise linear function, where each piece function ranges from statistic minimum to statistic maximum. In this way, we can flexibly modulate frame-wise reference motion as guidance. The detailed process is shown in Alg. \ref{alg:optimize}

\begin{algorithm}[H]\label{alg:optimize}
\SetAlgoLined
\caption{Optimization Process}
\KwIn{Noisy latents $z$ at timestep $t$, region mask $\mathcal{R}$, reference MOFT $\mathcal{M}^r$, the network $\mathcal{N}$, Motion Channel Mask $\mathcal{C}$, learning rate $\eta$}
\KwOut{Optimized latents $\hat{z}$}

\Begin{
    Get intermediate feature $\mathcal{X}$ from the network $\mathcal{N}$\;

    Given $\mathcal{X}$, $\mathcal{C}$, extract MOFT $\hat{\mathcal{M}}$ by Eq. \ref{alg:moft_extraction}\;

    Given $\mathcal{M}^r$, $\mathcal{M}$, and $\mathcal{R}$, compute the loss $\mathcal{L}$ by Eq. \ref{equ:motion_control_guidance}\;

    Optimize $\hat{z}$ by updating $\hat{z} \leftarrow z - \eta \nabla \mathcal{L}$\;

    \Return $\hat{z}$\;
}
\end{algorithm}

\begin{figure}[t]
    \centering
    \includegraphics[width=\linewidth]{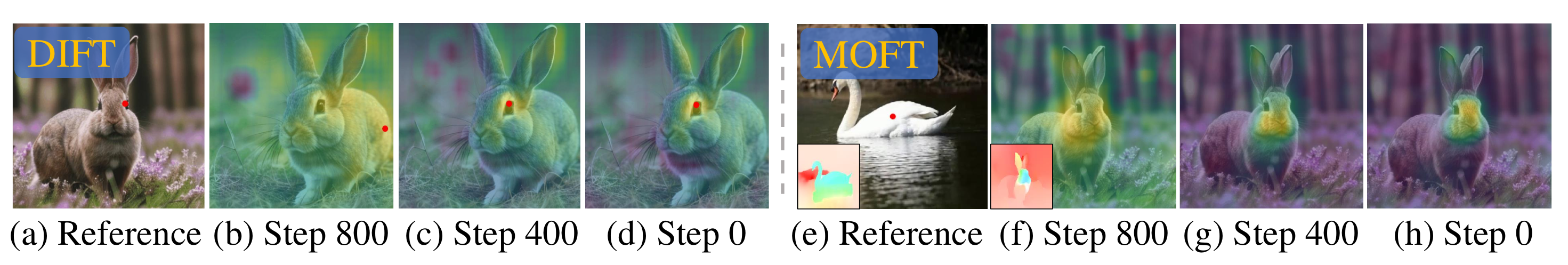}
    \vspace{-18pt}
    \caption{\textbf{Effects of DIFT and MOFT on different denoising time steps.} Given the source point in (a) (for DIFT) and (e) (for MOFT), we plot the similarity heat map of DIFT (b-d) and MOFT (f-h) of different denoising steps. Yellow indicates higher similarity. The \textcolor{red}{red} point in (b-d) indicates the position with highest similarity. It suggests that MOFT can provide more valid information than DIFT at the early denoising stages.}
    \vspace{-6pt}
    \label{fig:MOFT_vs_DIFT}
\end{figure}

\subsection{Point-Drag Manipulation} \label{point_drag_manipualtion}
Point-drag manipulation is designed to precisely relocate points within image and video frames to reach specific target points. In the image domain, this manipulation method often relies on motion supervision and point-tracking \cite{pan2023drag,shi2023dragdiffusion}, ensuring the precise tracking of point trajectories to achieve the desired target points. In the video domain, however, we can directly optimize whole point trajectories by setting targets in each frame. 
The loss function for optimizing point trajectories $\tau = {p_1, p_2, ... p_F}$ is:
\begin{equation}
\mathcal{L}^p = \sum_{i=2}^{F} ||\mathcal{D}(p_i) - \mathrm{sg}(\mathcal{D}(p_1))||,
\end{equation}
where $\mathcal{D}$ is the diffusion feature (DIFT) and $\mathrm{sg}$ is the "stop gradient" operation.

However, direct application of this method results in poor video motion control because DIFT struggles with semantic correspondence 
at early denoising steps, as shown in Fig. \ref{fig:MOFT_vs_DIFT} Row 1. Since spatial and temporal structures are already determined at early steps, DIFT's effectiveness is limited.
Conversely, MOFT provides relatively distinguished motion information in early denoising stage performance (Fig. \ref{fig:MOFT_vs_DIFT} Row 2), suggesting a strategy of using MOFT for initial coarse motion control and DIFT for precise point-drag manipulation. Please refer to Supplementary Material for details.






%% file: sections/experiments.tex
\section{Experiments}

\begin{figure}[t!]
    \centering
    \includegraphics[width=\linewidth]{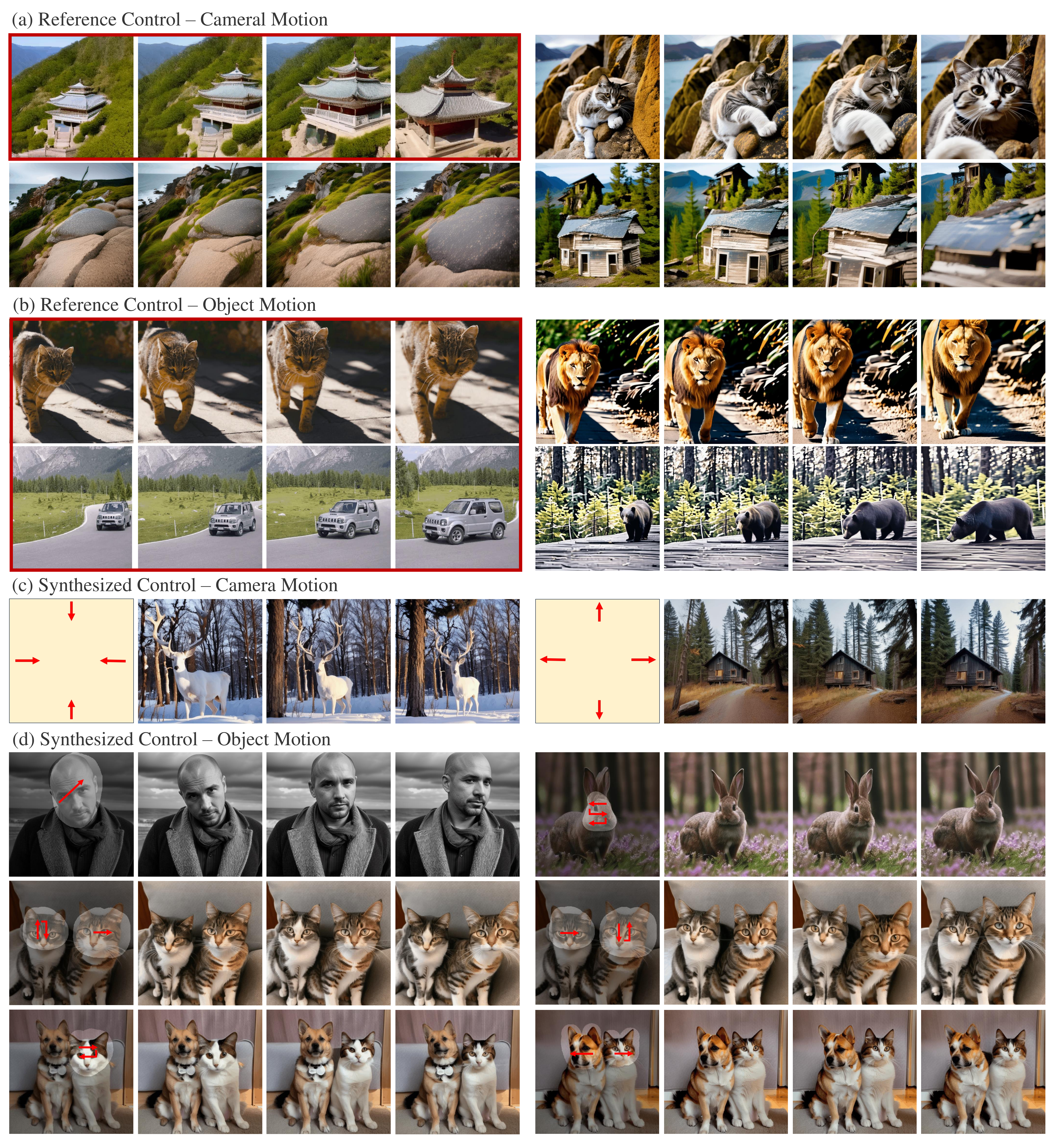}
    \vspace{-16pt}
    \caption{\textbf{Qualitative results.} We illustrate several animation clips with different reference or synthesized motion control signals. The \textcolor{red}{red} boxes in (a-b) stand for reference videos. \textit{We highly recommend readers refer to the supplementary material for a better visual experience.}}
    \vspace{-5pt}
    \label{fig:qualitative}
\end{figure}

\begin{figure}[t]
    \centering
    \includegraphics[width=\linewidth]{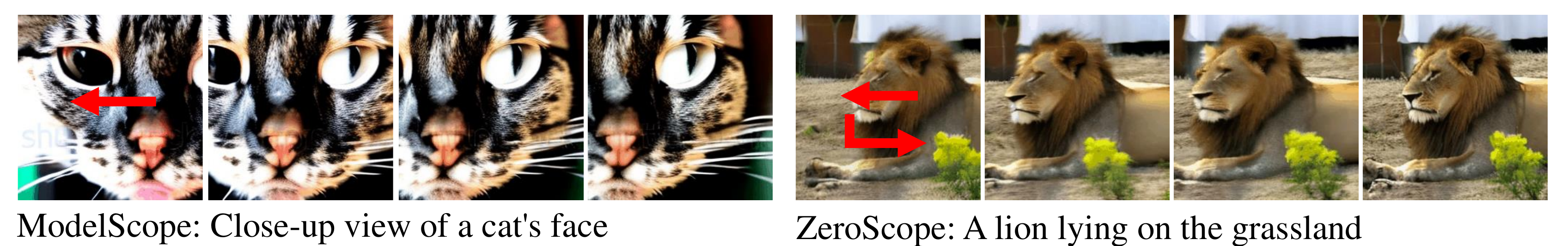}
    \vspace{-15pt}
    \caption{\textbf{Qualitative results on Modelscope \cite{wang2023modelscope} and ZeroScope \cite{zeroscope}.} }
    \vspace{-5pt}
    \label{fig:other_models}
\end{figure}

\subsection{Implementation details}
If not specified, the default video generation models of the following experiments are implemented in AnimateDiff \cite{guo2023animatediff} (T2V) and SparseCtrl \cite{guo2023sparsectrl} (I2V). For T2V generation, we first generate a normal video as the editing source, then apply motion direction and region mask to the video for motion control. To preserve consistency with the source video, we apply (1) region gradient clip, and (2) shared key and value. Details of these techniques and video results can be found in the Supplementary Material. Our results are at a resolution of 512x512 and 16 frames unless otherwise specified. We use DDIM with 25 denoising steps for each sample. It takes approximately 3 minutes to generate one sample on an RTX 3090 GPU. 

\subsection{Qualitative Results}
\begin{figure}[t!]
    \centering
    \includegraphics[width=0.80\linewidth]{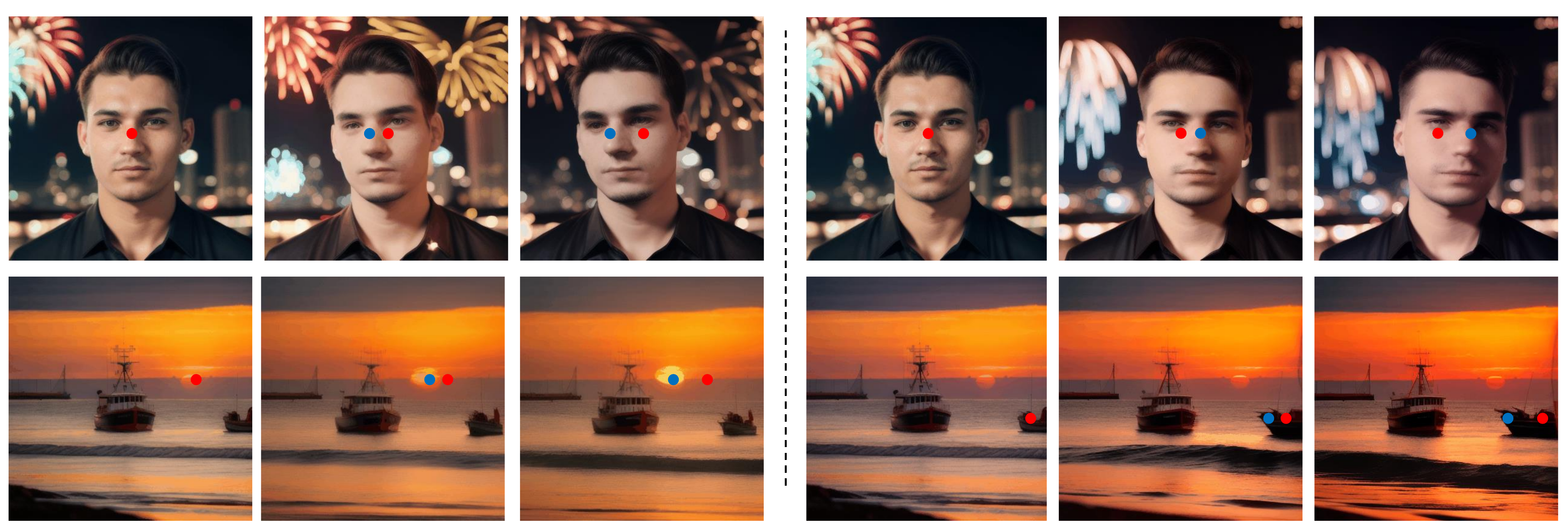}
    \vspace{-9pt}
    \caption{\textbf{Qualitative results of point-drag manipulation.} \textcolor{red}{Red} points indicate starting points. \textcolor{blue}{Blue} points indicate target points of the corresponding frames. We display three frames per video clip. 
    }
    \label{fig:drag}
\end{figure}

\vspace{-0.15cm}
We showcase qualitative outcomes in Fig.~\ref{fig:qualitative}. The figure illustrates the successful animation of videos by our method, guided by diverse control signals while preserving a natural and authentic sense of motion. Additionally, we exhibit the results of applying our motion control technique to alternative video generation models, such as ModelScope \cite{wang2023modelscope} and ZeroScope \cite{zeroscope}, employing the same control strategy (see Fig.~\ref{fig:other_models}). These results highlight the generalizability of MOFT across various video generation models.
We also showcase the application of our method on point-drag manipulation (Sec.~\ref{point_drag_manipualtion}) in Fig.~\ref{fig:drag}, where we successfully move the starting points to the targets.

\begin{table}[t]
  \small
  \caption{Experiments on Motion Feature Design
  }
  \vspace{-6pt}
  \label{tab:motion_feature_design}
  \centering
  \resizebox{0.75\textwidth}{!}{
  \begin{tabular}{c|c|c|c}
    \toprule
    Guide type & Methods & Motion Fidelity ($\uparrow$) & Imaging Quality ($\uparrow$)\\
    \midrule
    None & Origin & - & \textbf{0.697} \\
    \midrule
    \multirow{4}*{\makecell{Reference \\Guidance}}
    & SMM feature \cite{yatim2023space} & 70.2 & 0.681 \\
    & Vanilla Feature & 31.1 & 0.512 \\
    & + CR & 67.1 & 0.671\\
    & + CR \& MCF (Ours) & \underline{82.5} & \underline{0.693}\\
    \midrule
    \makecell{Synthesis \\Guidance}
    & Ours & \textbf{84.0} & \underline{0.693} \\
  \bottomrule
  \end{tabular}
  }
  \vspace{-0.3cm}
\end{table}

\subsection{Motion Feature Design}
\vspace{-0.15cm}
This subsection experiments on the effectiveness of motion feature designs with two metrics:
\textit{a) Motion Fidelity.}
Following \cite{yatim2023space}, we use Motion Fidelity to assess the fidelity of our results in the alignment of synthesis guidance or reference guidance. We use off-the-shelf tracking method \cite{karaev2023cotracker} to estimate the tracklets $\mathcal{T}=\{\tau_1, ..., \tau_M\}$ in the generated videos. For guidance, we manually construct synthesized tracklets for synthesis guidance and use estimated tracklets for reference guidance, we denote them both as $\mathcal{\widetilde{T}}=\{\widetilde{\tau}_1, ..., \widetilde{\tau}_N\}$ for simplicity. The motion fidelity score is defined as follows:
\begin{equation}
\frac{1}{m}\sum_{\widetilde{\tau}\in\mathcal{\widetilde{T}}}\max_{\tau\in\mathcal{T}}\mathbf{corr}(\tau,\widetilde{\tau})+\frac{1}{n}\sum_{\tau\in\mathcal{T}}\max_{\widetilde{\tau}\in\mathcal{\widetilde{T}}}\mathbf{corr}(\tau,\widetilde{\tau}).
\end{equation}
The correlation between two tracklets $\mathbf{corr}(\tau,\widetilde{\tau})$ is computed as:
\begin{equation}
    \mathbf{corr}(\tau,\widetilde{\tau}) = \frac{1}{F}\sum^F_{k=1}\frac{v^x_k\cdot\widetilde{v}^x_k+v^y_k\cdot\widetilde{v}^y_k}{\sqrt{(v^x_k)^2+(v^y_k)^2}\cdot\sqrt{(\widetilde{v}^x_k)^2+(\widetilde{v}^y_k)^2}},
\end{equation}
where $(v^x_k,v^y_k)$, $(\widetilde{v}^x_k,\widetilde{v}^y_k)$ are the $k^{th}$ frame displacement of tracklets $\tau$, $\widetilde{\tau}$, respectively.
\textit{b) Image Quality.} We follow \cite{huang2023vbench, ke2021musiq} that uses an image quality predictor trained on the SPAQ  dataset~\cite{fang2020perceptual} to evaluate frame-wise quality regarding distortion like noise, blur, or over-exposure. We collect a total of 270 prompt-motion direction pairs for experiments.

Table \ref{tab:motion_feature_design} summarizes our results. The vanilla feature shows poor motion fidelity and image quality due to extraneous information disrupting motion control. Removing content correlation significantly improves both metrics, yielding results comparable to the Space-Motion Map (SMM) feature \cite{yatim2023space}, likely because SMM also removes content correlation through frame-wise differences. MOFT guidance achieves the highest motion fidelity, with only a minor loss in image quality compared to the original unguided generation.



\begin{figure}[t!]
    \centering
    \includegraphics[width=0.80\linewidth]{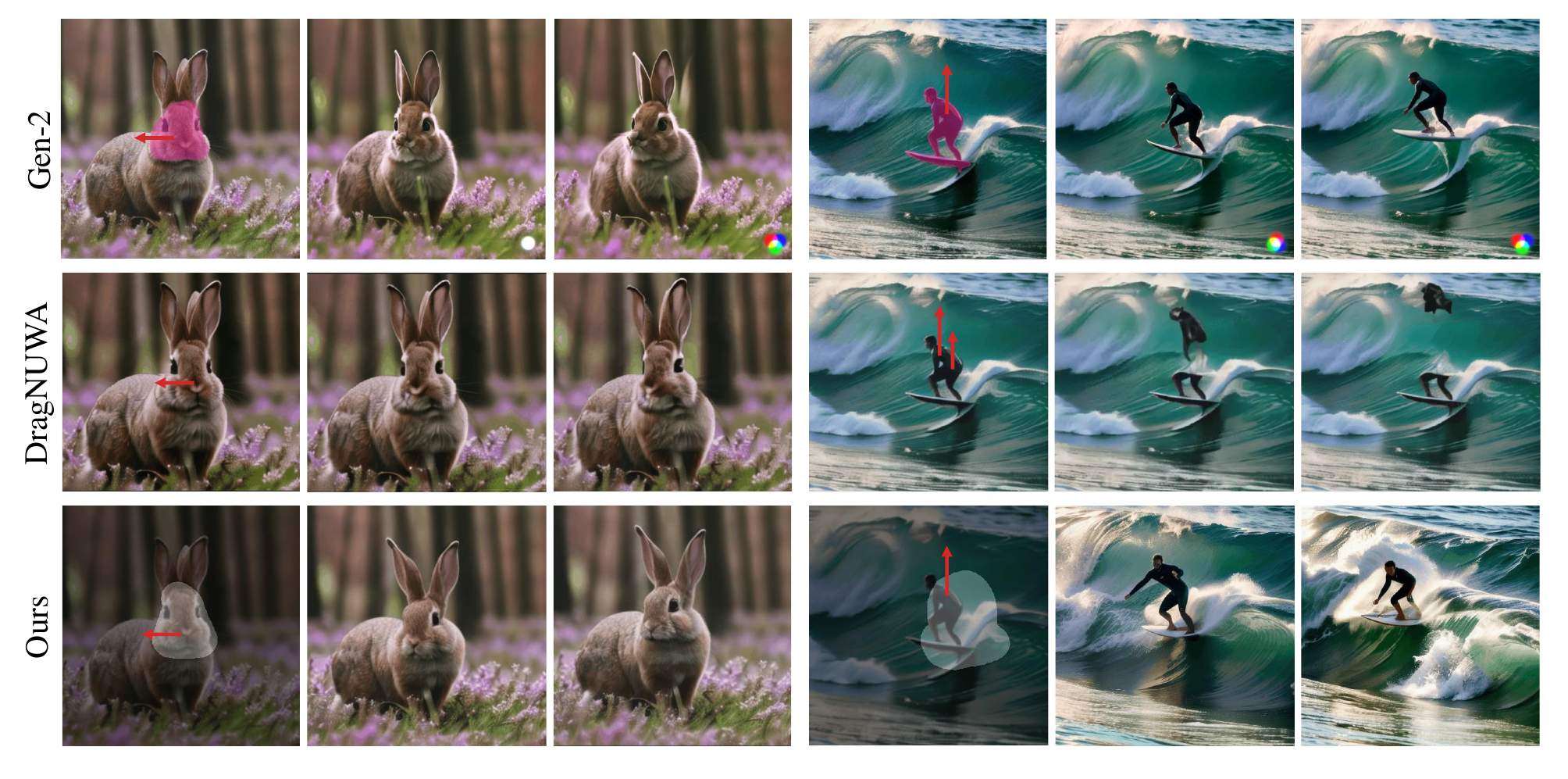}
    \vspace{-12pt}
    \caption{\textbf{Motion quality comparisons.} Gen-2 \cite{gen2} and Ours accept editing region and motion direction as the control signal. DragNUWA \cite{yin2023dragnuwa} accepts point trajectories as the control signal.}
    \label{fig:motion_quality_compare}
\end{figure}

\begin{table}[t!]
\hspace{12pt}
\begin{minipage}[t!]{0.37\textwidth}
\makeatletter\def\@captype{table}
\vspace{6pt}
\caption{Drag Precision}
\vspace{-4pt}
\scalebox{0.9}{
\begin{tabular}{cc}
    \toprule
    Name     &  Mean Distance ($\downarrow$)\\
    \midrule
    DragNUWA \cite{yin2023dragnuwa} &  \textbf{0.075} \\
    DIFT \cite{tang2024emergent} & 0.437\\
    + MOFT (Ours) & \underline{0.175} \\
    \bottomrule
\end{tabular}}
\label{point-drag}
\end{minipage}
\hspace{18pt}
\begin{minipage}[t!]{0.5\textwidth}
\makeatletter\def\@captype{table}
\vspace{6pt}
\caption{User Preference}
\vspace{-4pt}
\scalebox{0.9}{
\begin{tabular}{ccc}
    \toprule
    Methods     & Faithfulness ($\uparrow$) & Naturalness ($\uparrow$)\\
    \midrule  
    DragNUWA \cite{yin2023dragnuwa} &  2.50 & 2.08  \\
    Gen-2 MB \cite{gen2}  & \textbf{3.37} & \underline{2.90} \\
    Ours   &  \underline{3.21} &  \textbf{3.49}  \\
    \bottomrule
\end{tabular}}
\label{tab:user_study}
\end{minipage}
\vspace{-12pt}
\end{table}

\subsection{Point-drag Manipulation}
\vspace{-0.15cm}
We conducted additional experiments to assess the efficacy of incorporating motion control in point-based manipulation. In this comparison, we introduce DragNUWA \cite{yin2023dragnuwa}, a potent data-driven method, for reference. We follow \cite{pan2023drag, shi2023dragdiffusion} to use the \textit{Mean Distance} between edited points and target points to evaluate the drag precision. Specifically, we still use \cite{karaev2023cotracker} to estimate the tracklets $\mathcal{T}=\{\tau_1, ..., \tau_{,=M}\}$ of given small region. We average these tracklets into $\overline{\tau}$ and calculate the mean distance with target tracklet $\tau^{t}$. We normalize the final distance into [0,1], with 0 indicating no mean distance error. We collect a total of 40 image-motion direction pairs for experiments. As indicated in Table \ref{point-drag}, applying only DIFT guidance results in poor drag precision. By comparison, incorporating our MOFT yields substantial improvements, effectively narrowing the performance gap with the training-based DragNUWA. The finding is coherent with our analysis in Sec. \ref{point_drag_manipualtion}.

\subsection{User study}
\vspace{-0.15cm}
We conducted a survey to investigate users' preferences regarding videos generated with motion control. Employing a blind rating protocol, participants were randomly exposed to videos generated by Gen-2 Motion Brush \cite{gen2}, DragNUWA \cite{yin2023dragnuwa}, and our proposed method. Participants were instructed to rate from 1 to 5 (worst to best) on two metrics:
\textit{1) Motion Faithfulness} to measure how well the motion aligns with the control signal.
\textit{2) Motion Naturalness} to evaluate the naturalness and realism of the motion. We collect human feedback from 26 people on 56 video clips. As depicted in Table \ref{tab:user_study} and Fig. \ref{fig:motion_quality_compare}, it is evident that Gen-2 MB excels in achieving highly faithful motion control at the cost of motion naturalness. Gen-2 MB and DragNUWA tend to generate stiff and unrealistic motions.  In contrast, our proposed methods demonstrate competitive motion faithfulness while simultaneously preserving the natural and authentic quality of motion.

%% file: sections/conclusion.tex
\section{Limitations add Future Works}

While our approach has yielded appealing results, some limitations require future studies: 

1) Presently, our approach lacks support for motion control in real videos. Primarily, this limitation stems from the lack of research on video inversion techniques over video diffusion models. We have observed significant alterations in content when employing initial noise from DDIM inversion~\cite{song2020denoising} on real videos. Future research focused on video inversion holds promise for resolving this issue.

2) Our current approach does not allow for precise motion scale guidance in motion control. While there are strategies to roughly control motion scales, such as adding up control weights for larger motion scales or implementing gradient clips for smaller ones, achieving high precision in motion scale manipulation requires further investigation.

\section{Conclusion}
\vspace{-0.15cm}
In summary, our analysis reveals a robust motion-aware feature in video diffusion models, leading to the development of a training-free MOtion FeaTure (MOFT). MOFT encodes rich, interpretable motion information, is extracted without training, and is applicable across diverse architectures. We introduce a novel training-free video motion control framework based on MOFT, demonstrating competitive performance with natural motion. Importantly, our approach is versatile, easily adaptable to various architectures and checkpoints without the need for independent training.

\myparagraph{Acknowledgements.} This research is supported by MOE AcRF Tier 1 (RG97/23) and is also supported under the RIE2020 Industry Alignment Fund – Industry Collaboration Projects (IAF-ICP) Funding Initiative, as well as cash and in-kind contribution from the industry partner(s).

%% file: sections/supp.tex
\section{Supplementory Material}

In this section, we provide more analysis and results. It is highly recommended to refer to the attached webpage for better visual illustrations.




\subsection{Preliminary for Latent Optimization}
Diffusion models learn to recover clean images $x$ from random noise $z_T\sim\mathcal{N}(0,I)$ with a sequential denoising process \cite{ho2020denoising, sohl2015deep, song2020score}. \cite{rombach2022high} proposed the latent diffusion model (LDM), which maps data into a lower-dimensional space via a variational auto-encoder (VAE) \cite{kingma2013auto} and models the distribution of the latent embeddings instead.
At the diffusion step $t$, random noise $\epsilon_t$ is added to $x$, giving a noisy image $z_t = \alpha_t x+\sigma_t\epsilon_t$, with $\alpha_t$ and $\sigma_t$ the time-dependent parameters.
The estimation of the denoised image is equivalent to predicting the noise $\epsilon_t$.

Our latent optimization strategy is motivated by \cite{shi2023dragdiffusion, yatim2023space} that uses intermediate features to supervise the latent optimization process, which can be formulated as
\begin{equation}
    z^{new}_t = z_t - \eta \frac{\partial \mathcal{L}}{\partial z_t},
\end{equation}
where $\eta$ is the learning rate and $\mathcal{L}$ is the loss function.

\subsection{More Analysis and Details}

\begin{figure}[h!]
    \centering
    \includegraphics[width=0.80\linewidth]{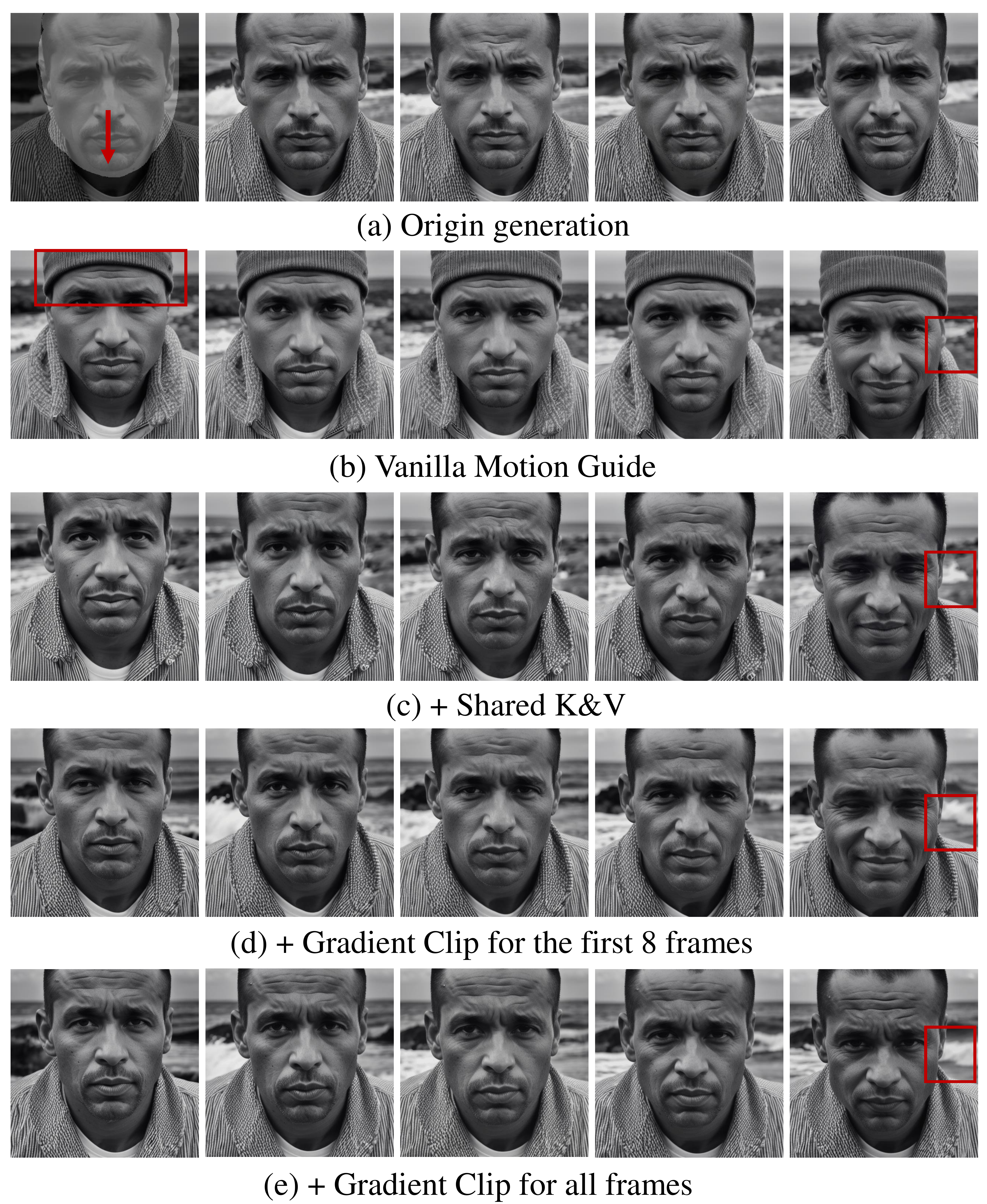}
    \caption{\textbf{Qualitative comparison on video consistency preservation.} We compare the generated results w./wo. our introduced techniques. The control signal is shown in the first image of (a), with the red arrow indicating the motion control direction and the light region indicating the control region. We highlight the noticeable region with red boxes. It reveals that Shared K\&V contributes to the consistency of the whole video. Gradient Clip adds consistency out of masked regions but meanwhile reduces motion scale.}
    \label{fig:consistency_ablation}
\end{figure}

\myparagraph{Video Consistency Preservation.}
Since one of our applications is to control the motion in source-generated videos, it is important to preserve the consistency between the source video and the target video. To this end, we introduce two techniques: Shared K\&V and Masked Gradient Clip. We visualize their qualitative comparison in Fig. \ref{fig:consistency_ablation}.

\begin{figure}[h]
    \centering
    \includegraphics[width=0.6\linewidth]{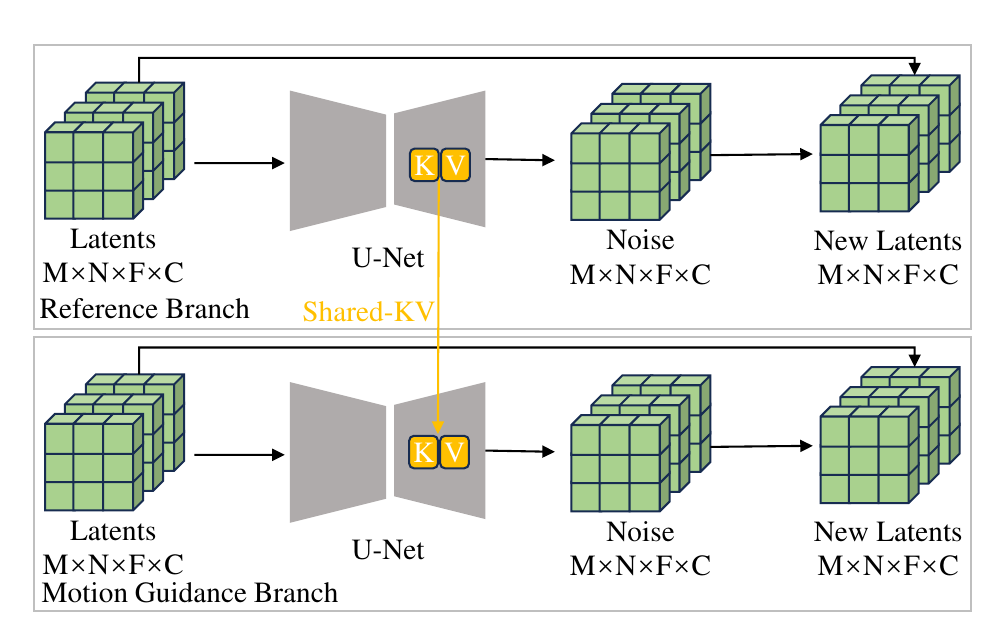}
    \caption{\textbf{Motion Control Pipeline with Shared K\&V.} We apply the origin designing process in the reference branch while applying motion guidance in the motion guidance branch. During denoising, we insert the K\&V of the reference branch to the motion guidance branch for content preservation.}
    \label{fig:shared_kv}
\end{figure}

\textit{Shared K\&V.}
As proved by many previous works \cite{shi2023dragdiffusion,mou2023dragondiffusion,cao2023masactrl}, inserting Key (K) \& Value (V) of spatial attention from reference branch to target branch can help to preserve content information of reference generation. As shown in Fig. \ref{fig:shared_kv}, we adopt this method to our motion control pipeline. 

As shown in Fig. \ref{fig:consistency_ablation} (a-c), shared K\&V contributes to the consistency of the whole video. The generated video with vanilla motion guidance (Fig. \ref{fig:consistency_ablation} (b)) adds additional contents (i.e. a hat on the man's head) while adding shared K\&V (Fig. \ref{fig:consistency_ablation} (c)) stays consistent with the original generation.

\textit{Masked Gradient Clip.} Since we do not want to change much content out of the masked region during motion guidance optimization, we simply clip the guidance gradient $g$ out of the masked region, which is
\begin{equation}
g^{\text{clip}}= 
\begin{cases}
g, & (i,j,k)\in\mathcal{R}\&k \in \mathcal{F} \\
0, & \text{else}, \\
\end{cases}
\label{eqn:gradient_clip}
\end{equation}
where $i,j,k$ are indices of height, width, and frame, respectively. $\mathcal{R}$ is the spatial mask region index set. $\mathcal{F}$ is the frame set. As shown in Fig. \ref{fig:consistency_ablation} (d-e), gradient clipping adds consistency to the background content. While applying gradient clipping to more frames increases consistency, it also results in a smaller motion scale. Thus, applying gradient clipping involves a trade-off. In practice, we apply gradient clipping to the first 8 frames.

\begin{figure}[h!]
    \centering
    \includegraphics[width=0.5\textwidth]{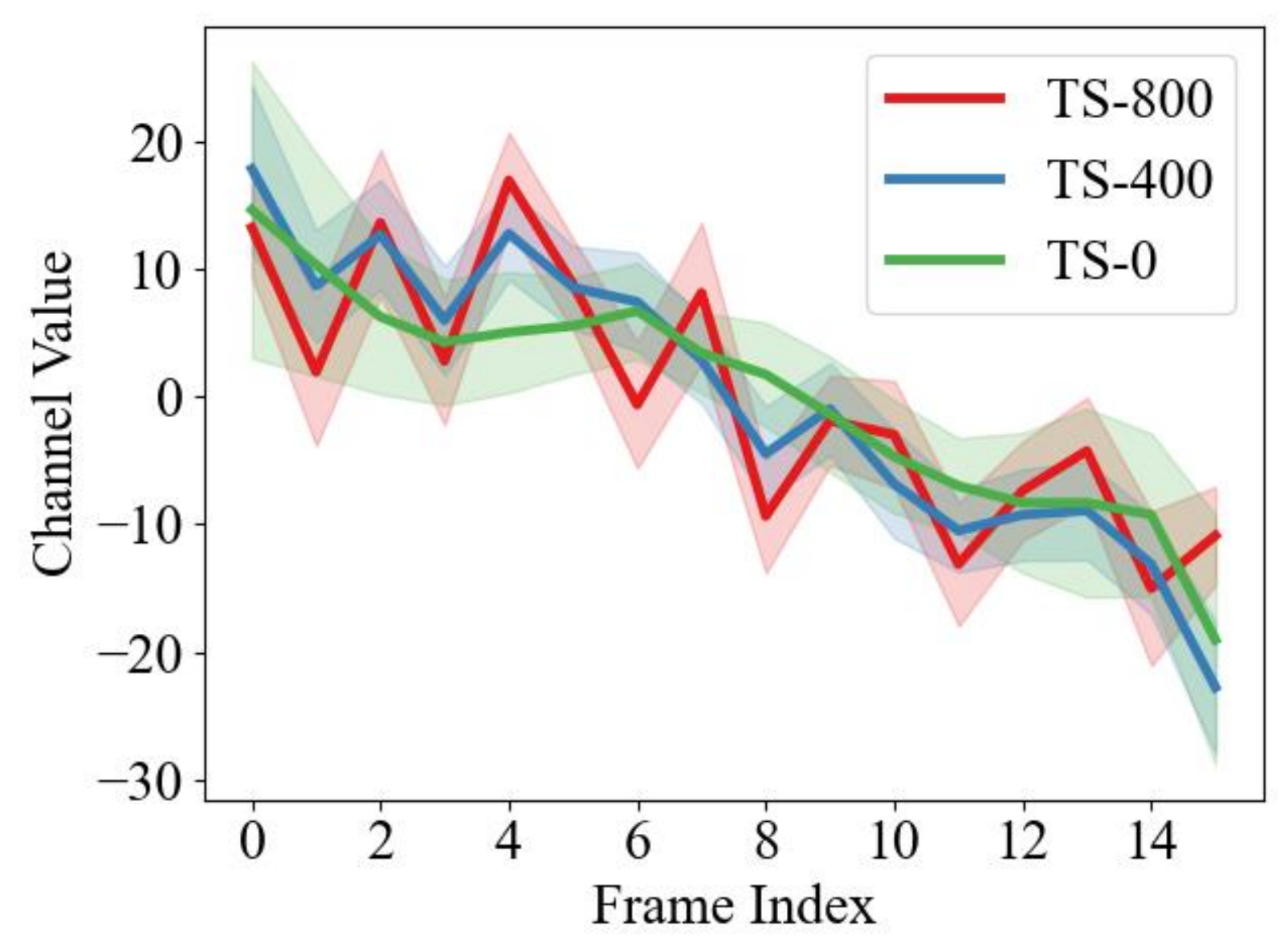}\vspace{-2mm}
    \caption{\textbf{Motion channel value for different denoising timestep.} TS-800 indicates denoising step 800.}
    \vspace{-12pt}
    \label{fig:line_char_diff_timestep}
\end{figure}

\myparagraph{Timestep Choice for Motion Extracted from Video.}
As shown in Fig. \ref{fig:line_char_diff_timestep}, the trends and ranges of motion channels with the same motion direction are similar among different denoising timesteps, while the curve nearing the end of denoising steps is smoother and refined. To this end, we use the MOFT at the beginning of the inversion stage as the guidance for all control timesteps. 

 
            

\myparagraph{Motion Channel Filter Number.}
We ablate the effects to filter different numbers of motion channels in Tab. \ref{tab:channel_ablation}. 
\begin{table}[t]
  \caption{Motion Channel Ablation.}
  \label{tab:channel_ablation}
  \centering
  \vspace{-6pt}
  \begin{tabular}{c|cc}
    \toprule
    Channel number     & Faithfulness ($\uparrow$) & Naturalness ($\uparrow$)\\
    \midrule  
    top 100\% & 67.5 & 0.671 \\
    top 50\% &70.5 & 0.681\\
    top 10\% & 82.7 & 0.692\\
    top 5\% & 83.2 & \textbf{0.694} \\
    top 4\% & \textbf{84.0} & \underline{0.693} \\
    top 3\% & \underline{83.6} & \underline{0.693} \\
    top 1\% & 76.3 & 0.677\\
  \bottomrule
  \end{tabular}
  \vspace{-0.3cm}
\end{table}
The key finding is that preserving only a few channels most sensitive to motion can enhance both motion faithfulness and naturalness, as the effects of irrelevant information in other channels are removed.
However, when we further reduce the channel number to the top 1\%, both motion faithfulness and naturalness significantly decrease due to the loss of some motion-sensitive channels. 
In practice, we choose the top 4\% of motion channels.

\begin{figure}[h!]
    \centering
    \includegraphics[width=0.4\linewidth]{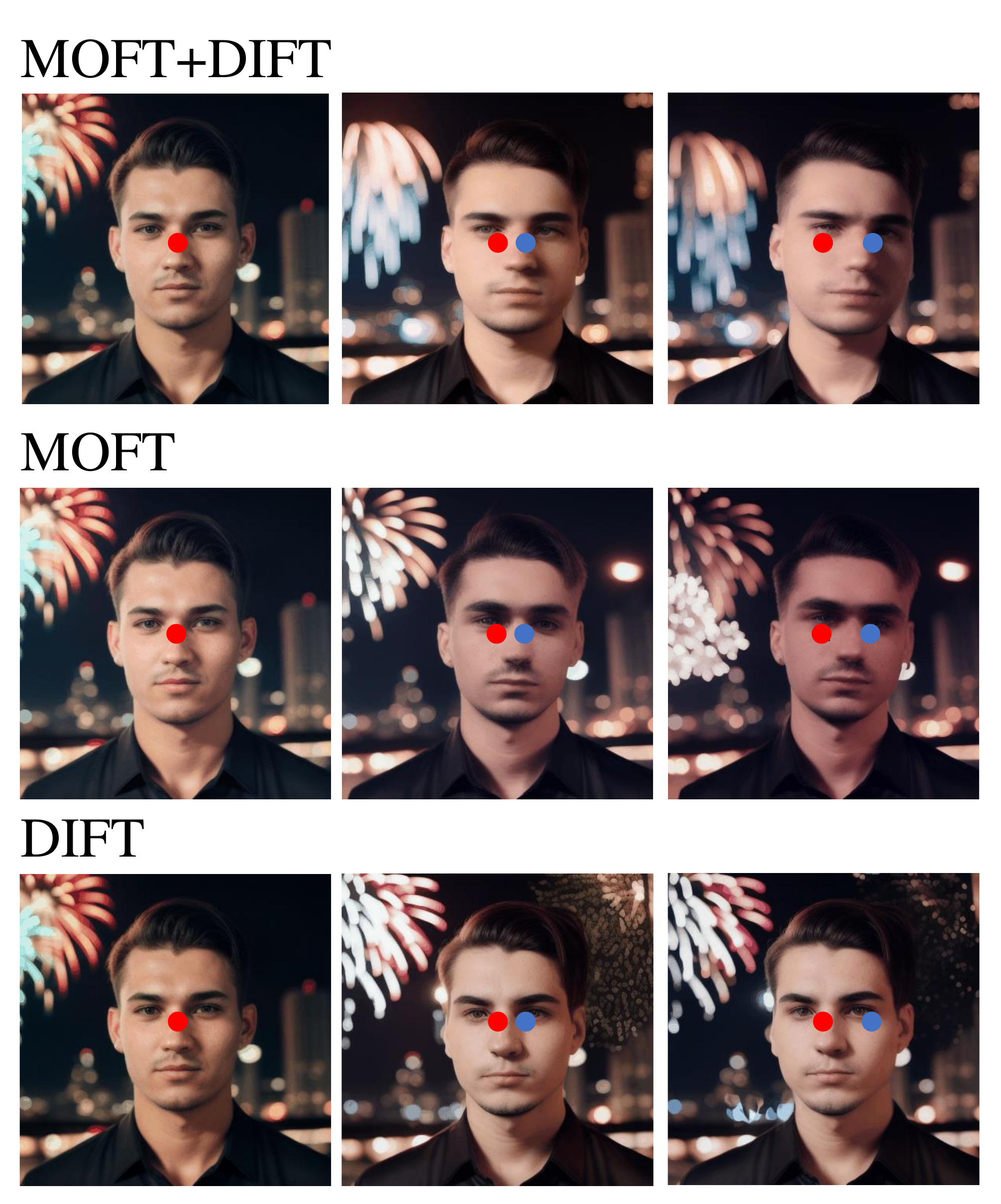}
    \caption{\textbf{Qualitative results of point-drag manipulation ablation.} Red points indicate starting points. Blue points indicate ending points. We only display three frames per animation clip. 
    }
    \label{fig:drag_ablation}
\end{figure}

\myparagraph{Point-Drag Manipulation Ablation.} Following the main paper Sec. 4.2, we use MOFT for initial coarse motion control and DIFT for precise point-drag manipulation. The compositional loss function is
\begin{equation}
\mathcal{L}_t= w^c_t \mathcal{L}^c + w^p_t \mathcal{L}^p, 
\begin{cases}
w^c_t >0, w^p_t=0, & \mathrm{if}\quad t >= t_1 \\
w^c_t >0, w^p_t>0, & \mathrm{if}\quad t_1 > t >= t_2 \\
w^c_t =0, w^p_t>0, & \mathrm{if}\quad t_2 > t >= t_3 \\
w^c_t =0, w^p_t=0, & \mathrm{if}\quad t_3 > t \\
\end{cases}
\label{eqn:baseline_objective}
\end{equation}
where $w^c_t$ and $w^p_t$ are time-dependent weights under the threshold $t_1$, $t_2$ and $t_3$. In practice, the total denoising step is 25. We set $t_1=19$, $t_2=18$, $t_3=5$. We further ablate the effectiveness of the design in Fig. \ref{fig:drag_ablation}. Applying only DIFT results in limited motion. Using only MOFT produces motion but lacks precise point control. By combining DIFT and MOFT, we achieve precise point-drag control.

\subsection{More Visualization}

\begin{figure}[h]
    \centering
    \includegraphics[width=\linewidth]{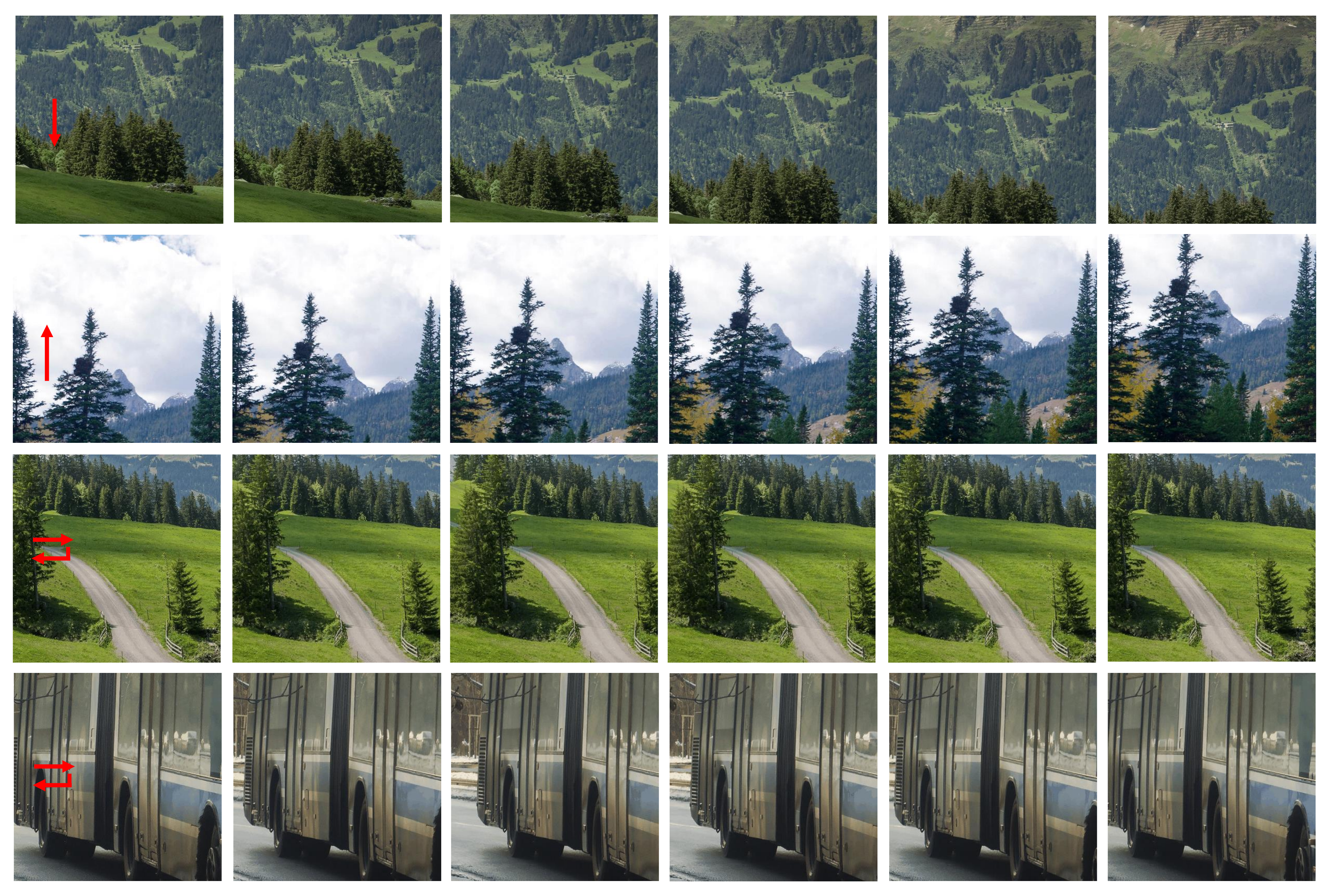}
    \caption{\textbf{Videos for PCA.} We manually move the whole picture following specified motion directions to synthesis videos.}
    \label{fig:motion_direction_example}
\end{figure}

\myparagraph{PCA video examples.}
In Fig. \ref{fig:motion_direction_example}, we provide some of the video examples that we use to conduct PCA. We manually move the whole picture following specified motion directions to synthesis videos.

\begin{figure}[h]
    \centering
    \includegraphics[width=\linewidth]{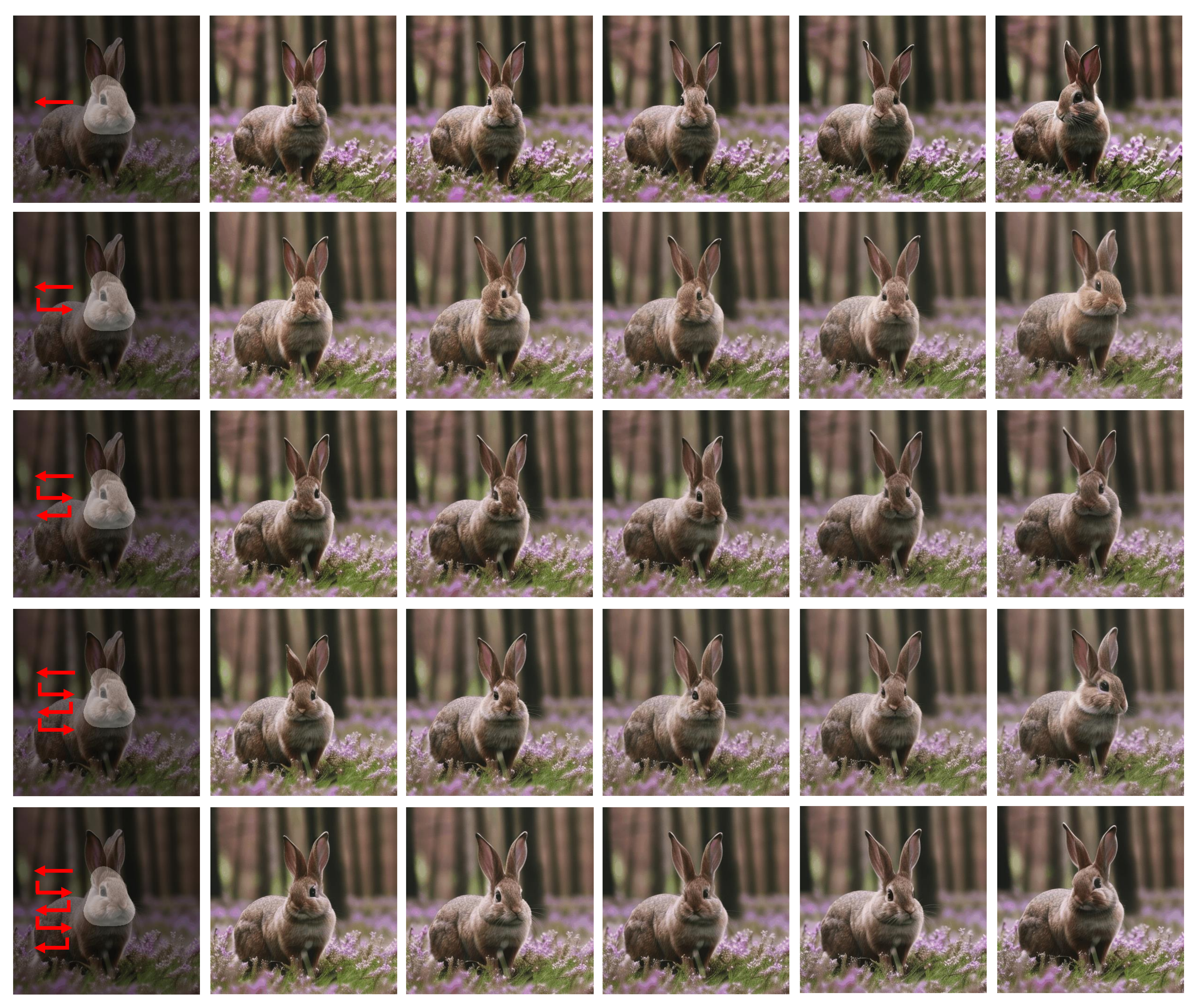}
    \caption{\textbf{More qualitative results.}}
    \label{fig:supp_qualitative_results}
\end{figure}

\myparagraph{Qualitative Results.}
We provide some qualitative results in Fig. \ref{fig:supp_qualitative_results}. More animated results can be found on the attached webpage.

\begin{figure}[h]
    \centering
    \includegraphics[width=\linewidth]{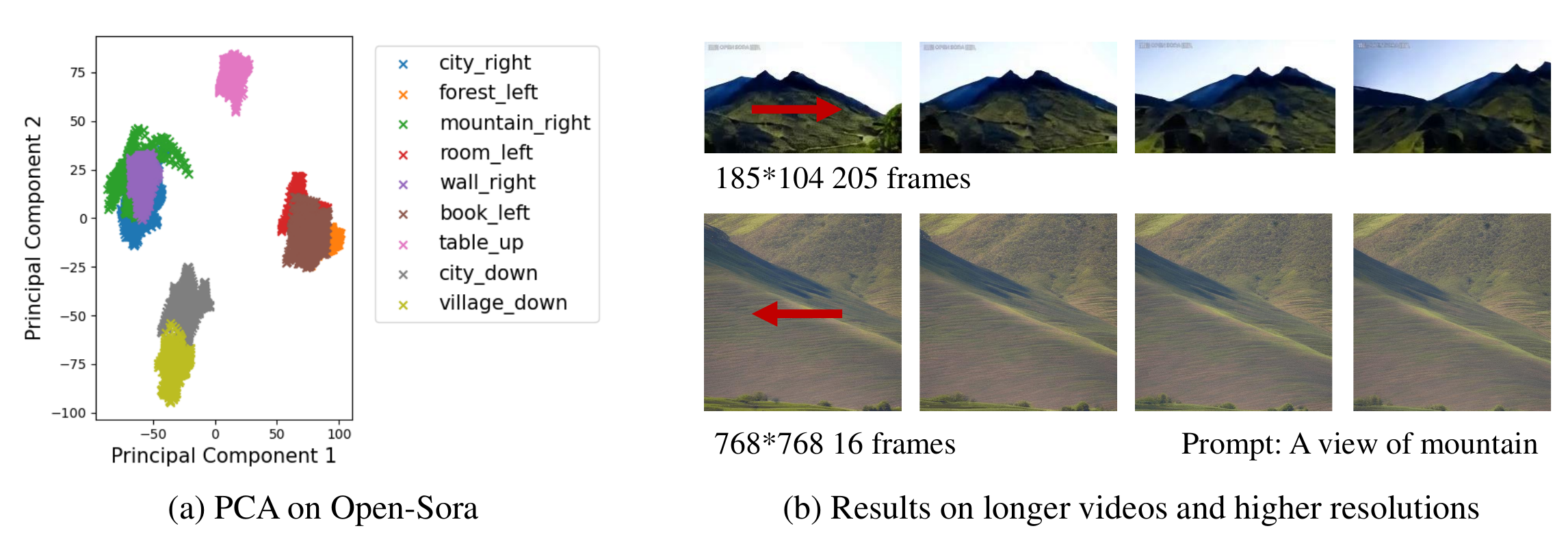}
    \caption{\textbf{More results on Open-Sora.}}
    \label{fig:rebuttal_opensora}
\end{figure}

\subsection{More Results on Open-Sora \cite{opensora}}
In Fig. \ref{fig:rebuttal_opensora}(a), we demonstrate that PCA can clearly separate videos with different motions based on their diffusion features from Open-Sora \cite{opensora}, an open-source video generation model capable of producing long videos. In Fig. \ref{fig:rebuttal_opensora}(b), we show that our methods can be applied to higher resolutions (768×768) and longer videos (205 frames on Open-Sora).
